%% file: main.tex
\documentclass{esannV2}
\usepackage[dvips]{graphicx}
\usepackage[latin1,utf8]{inputenc}
\usepackage{amssymb,amsmath,array}

\usepackage{adjustbox}
\usepackage{wrapfig}
\usepackage{graphicx}
\usepackage{amssymb,amsmath,array}
\usepackage{pifont}
\usepackage{multirow,multicol,makecell}
\usepackage{url}
\usepackage{latexsym}
\usepackage{array}
\usepackage{todonotes}
\usepackage{subfigure} 
\usepackage{algorithm}
\usepackage{algorithmic}
\usepackage{makecell}
\usepackage{booktabs}
\usepackage{listings}
\usepackage{csquotes}
\usepackage{amsthm}
\usepackage{bm}
\usepackage{xcolor,colortbl}
\usepackage{xspace}
\usepackage{cleveref}
\crefformat{section}{\S#2#1#3}
\crefformat{subsection}{\S#2#1#3}

\newcommand\tapt{\textsc{TAPT}\xspace}

\newcommand\pcp{\textsc{PCP}\xspace}
\newcommand\cls{\texttt{CLS}\xspace}

\newcommand\ft{\textsc{FT}\xspace}

\newcommand\nlp{\textsc{NLP}\xspace}
\newcommand\mlm{\textsc{MLM}\xspace}

\newcommand\robertalarge{\textsc{RoBERTa-}{\footnotesize \textsc{Large}}\xspace}

\newcommand\tf[1]{\textbf{#1}}
\newcommand{\tableindent}{~~}

\definecolor{legend1}{HTML}{66c2a4}
\definecolor{legend2}{HTML}{fa8c62}
\definecolor{legend3}{HTML}{8da0cb}
\definecolor{cid}{HTML}{dae8f5}
\definecolor{ccon}{HTML}{fee9d4}
\definecolor{gred}{HTML}{cc0200}
\definecolor{ggreen}{HTML}{4C9F26}
\definecolor{Gray}{gray}{0.93}
\newcommand\hl{\cellcolor{cid}}

\newcommand{\ua}{\textcolor{ggreen}{$\uparrow$}}
\newcommand{\da}{\textcolor{gred}{$\downarrow$}}

\newcommand{\ie}[0]{\emph{i.e., }}

\begin{document}
\title{Rethink the Effectiveness of Text Data Augmentation: An Empirical Analysis}

\author{Zhengxiang Shi and Aldo Lipani
%
%
\vspace{.3cm}\\
%
University College London \\
Gower St, London - United Kingdom
%
}

\maketitle

\begin{abstract}
\input{paper/abstract}

\end{abstract}

\section{Introduction}
\input{paper/introduction}

\section{Related Work}
\input{paper/related_work}

\section{Background}
\input{paper/method}
\section{Experiments}
\label{sec:experiment}
\input{paper/experiment}

\section{Conclusion}
\input{paper/conclusion}


\begin{footnotesize}
\bibliographystyle{unsrt}
\bibliography{main}
\end{footnotesize}


\end{document}

%% file: paper/abstract.tex
In recent years, language models (LMs) have made remarkable progress in advancing the field of natural language processing (\nlp). However, the impact of data augmentation (DA) techniques on the fine-tuning (FT) performance of these LMs has been a topic of ongoing debate. In this study, we evaluate the effectiveness of three different FT methods in conjugation with back-translation across an array of 7 diverse \nlp tasks, including classification and regression types, covering single-sentence and sentence-pair tasks.
Contrary to prior assumptions that DA does not contribute to the enhancement of LMs' FT performance, our findings reveal that continued pre-training on augmented data can effectively improve the FT performance of the downstream tasks. 
In the most favourable case, continued pre-training improves the performance of FT by more than 10\% in the few-shot learning setting.
Our finding highlights the potential of DA as a powerful tool for bolstering LMs' performance.\footnote{The code is available at \url{https://github.com/ZhengxiangShi/PowerfulPromptFT}.}

%% file: paper/introduction.tex
In recent years, the development of LMs has revolutionized the field of \nlp \cite{ni2020natural,ni2023}, leading to remarkable progress in a range of downstream tasks, including text classification \cite{Shi_Zhang_Lipani_2022,Shi2022attention}, information retrieval \cite{rahmani2020joint,fu2023,10.1145/3539813.3545126,shi2023self}, and multi-modalities \cite{shi-etal-2022-learning,10.1007/978-3-030-99736-6_20,shi2023and}. While LMs have shown impressive performance in many tasks, there has been a debate over the effectiveness of simple data augmentation (DA) techniques, such as back-translation, for improving the FT performance.

Previous research \cite{longpre-etal-2020-effective} evaluated DA techniques, such as Back-Translation, suggesting that these prevalent task-agnostic DA yields limited and inconsistent improvements for pre-trained LMs \cite{liu2019roberta} in many basic classification tasks.
Additionally, \cite{zhou-etal-2022-flipda} contended that most previous augmentation methods offer only marginal gains and are generally ineffective, pointing out that DA often leads to unstable performance and can trigger a failure mode, characterized by severe performance drops or fluctuations.

\begin{figure}[!t]
  \centering
  \includegraphics[width=\textwidth]{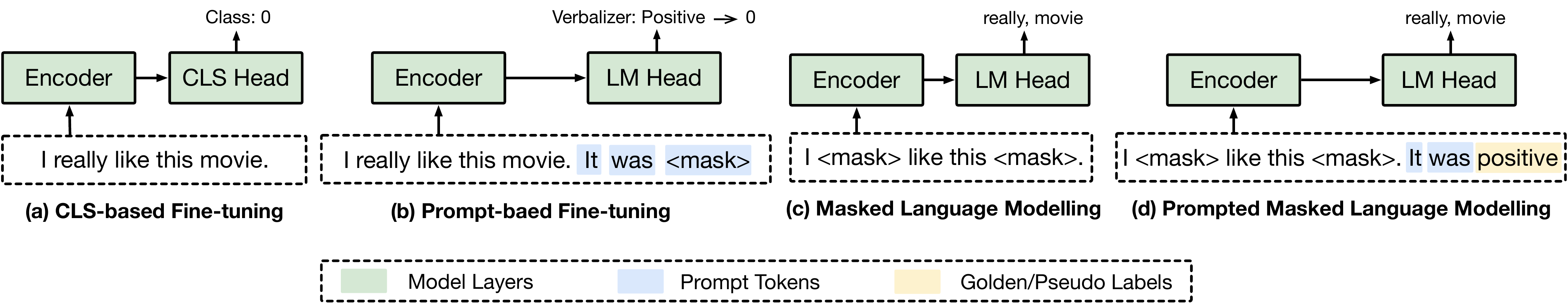}
  \caption{The overview of \cls-based and prompt-based \ft, along with their corresponding continued pre-training objectives.}
  \vspace{-1.2em}
  \label{fig:idea}
\end{figure}

In this study, we provide an empirical study to re-evaluate the effectiveness of text DA with two state-of-the-art prompt-based FT approaches \cite{gao-etal-2021-making,zhang2022differentiable}, as well as the conventional \cls-based FT \cite{liu2019roberta}, as shown in Figure \ref{fig:idea}(a,b). We perform experiments on seven distinct NLP tasks, including classification and regression tasks that involve single sentences and sentence pairs, to assess the efficacy of DA. Our findings contest the previously held belief that DA does not enhance LMs' FT performance. We discover that continued pre-training LMs on augmented data can largely improve the performance of FT approaches, offering an efficient alternative for enhancing model performance in practical applications.

%% file: paper/related_work.tex
\paragraph{\textbf{Prompt-based methods.}} In recent years, the exploration of prompt-based approaches has been conducted to enhance FT performance. PET/iPET \cite{schick-schutze-2021-exploiting} adapted the \cls-based \ft \cite{liu2019roberta} by presenting it as a masked language modelling problem, which is considered better suited for pre-training objectives, as illustrated in Figure \ref{fig:idea}. Subsequent studies further refined the application of templates and label words through automatic search mechanisms \cite{gao-etal-2021-making} or soft prompts that can be updated independently of any words \cite{zhang2022differentiable}.

\paragraph{\textbf{Continued Pre-training.}} Earlier research, such as \cite{longpre-etal-2020-effective,zhou-etal-2022-flipda}, has questioned the efficacy of simple DA in improving FT performance for downstream tasks. Prior studies \cite{shi2023rethinking,Shi2023} have demonstrated the effectiveness of continued pre-training on improving the model performance even with hundreds of unlabelled examples. However, the effectiveness of continued pre-training on the back translation augmented data is unclear in the context of few-shot learning.

%% file: paper/method.tex
In this section, we provide a brief overview of FT approaches and their respective continued pre-training methods. Figure \ref{fig:idea}(a) illustrates the conventional \cls-based \ft \cite{liu2019roberta}, which trains the output vector of the \texttt{[CLS]} token using an additional head layer. Further pre-training on task-related texts (see Figure \ref{fig:idea}c) before \cls-based \ft typically leads to improved model performance \cite{gururangan-etal-2020-dont,shi2023rethinking}. 

However, there exists a discrepancy between the pre-training objective and the \cls-based \ft objective, leading to prompt-based research to enhance language model performance. Figure \ref{fig:idea}(b) demonstrates that prompt-based \ft is designed as an \mlm problem with the objective of predicting the masked token \cite{schick-schutze-2021-exploiting}. Specifically, the input text $X$ is conditioned using a specific prompt template $\tilde{X} = \mathcal{T}(X)$, containing one special token, \texttt{[MASK]}. The prompt-based \ft then connects the output vector related to the \texttt{[MASK]} token to a label word. The probability of predicting class $y \in \mathcal{Y}$ is calculated as follows:

\begin{equation}
\label{eqn:marginal-prob}
p(y | X) = p(\textrm{\tt[MASK]} = \mathcal{M}(y) | \tilde{X}),
\end{equation}
where the verbalizer $\mathcal{M}: \mathcal{Y} \rightarrow \mathcal{V}$ maps the task label space to individual words in the vocabulary $\mathcal{V}$. Prompt-based \ft can employ either hard or soft prompt templates $\mathcal{T}$, with label words possibly being part of the prompt templates as well \cite{zhang2022differentiable}. Hard prompt templates \cite{schick-schutze-2021-exploiting} necessitate the careful design of prompts and label words for each task. However, the use of hard prompts was found to be sub-optimal and sensitive to prompt selection. Soft prompts \cite{zhang2022differentiable} were proposed to utilize unused tokens from the vocabulary $\mathcal{V}$ or additional tokens as tunable embeddings for prompt templates, which can be directly trained with task-specific supervision. A recent study \cite{Shi2023} proposed prompt-based continued pre-training prior to prompt-based \ft to further enhance language model performance on downstream tasks, as depicted in Figure \ref{fig:idea}(d).

%% file: paper/experiment.tex
In this section, we assess the impact of DA (\ie back translation) on all comparison methods. Additionally, we present datasets and baselines.

\input{table/dataset_statistics}
\paragraph{\textbf{Datasets.}}
Our study performs a comprehensive analysis of 7 \nlp datasets, including classification and regression tasks. 
We derive 6 single-sentence tasks (SST-5~\cite{socher2013recursive_sst-2}, MR~\cite{pang2005seeing_mr}, CR~\cite{hu2004mining_cr}, MPQA~\cite{wiebe2005annotating_mpqa}, Subj~\cite{pang2004sentimental_subj}, TREC~\cite{voorhees2000building_trec}) and 1 sentence-pair English tasks (STS-B~\cite{cer2017semeval_sts-b}), as shown in Table \ref{table:datasets}.
According to \cite{schick-schutze-2021-exploiting,gao-etal-2021-making,zhang2022differentiable,Shi2023}, we sample $K$-shot ($K$=16) per class from the full training set of each dataset.

\paragraph{\textbf{Baselines.}} 
We train the $K$-shot examples using three different FT approaches, either incorporating back-translation as DA or not. The approaches are as follows:
(1) ``\cls-based \ft'': see Figure \ref{fig:idea}a;
(2) ``Prompt-based \ft (hard)'': FT with high-quality manual or auto-generated prompts and label words \cite{schick-schutze-2021-exploiting} (see Figure \ref{fig:idea}b). Please refer to Table \ref{table:template} for the template details; and
(3) ``Prompt-based \ft (soft)'': FT with soft prompts using additional tokens for both templates and label words \cite{zhang2022differentiable}, where the same template is applied to all tasks (see Figure \ref{fig:idea}b). We use the SST-5 and STS-B templates for all single-sentence tasks and sentence pair tasks, respectively.

\input{table/template}

To compare the effectiveness of direct supervision learning on augmented data from back-translation \cite{ott2019fairseq}, we use augmented data as the corpus for continued pre-training with a masked language modelling objective. Consequently, we train these three types of \ft approaches from three different types of checkpoints to evaluate their relative effectiveness:
(i) the off-the-shelf \robertalarge checkpoint \cite{liu2019roberta};
(ii) the task-adaptive pre-training (\tapt) checkpoint \cite{gururangan-etal-2020-dont,chen-etal-2022-adaprompt} for \cls-based \ft; and
(iii) the prompt-based continued pre-training (\pcp) checkpoint \cite{Shi2023} for prompt-based \ft.

\paragraph{\textbf{Training Details.}}
We perform a grid search for learning rates within the set \{1e5, 2e-5, 5e-5\} with a batch size of 8. We train the model for 1,000 steps, evaluate performance every 100 steps, and select the best model based on the evaluation set.
We augment each example using English-German and English-Russian translations, resulting in two augmented examples per original example.

\input{table/results}

\vspace{-0.3em}
\paragraph{\textbf{Results.}}
Table \ref{table:few_shot} presents the performance of three different FT approaches, which involve using augmented examples as either supervised or continued pre-training training instances.
Our experimental results reveal two primary observations: 
(1) using augmented examples for continued pre-training (\tapt or \pcp) typically results in greater improvements compared to using them in supervised learning, and
(2) continued pre-training occasionally leads to considerable performance enhancements. We delve into these findings below.

\vspace{-0.6em}
\paragraph{\textbf{\#1. }} 
Continued pre-training (\tapt or \pcp) on three different FT approaches results in performance enhancements in 18 out of 21 cases, whereas using augmented data for supervised training leads to improvements in only 11 out of 21 cases. Furthermore, the average performance of FT with continued pre-training is 77.0\% across all datasets and FT approaches, while the average performance of FT using supervised training on augmented data is approximately 75.5\%. These results highlight the benefits of continued pre-training.

\vspace{-0.6em}
\paragraph{\textbf{\#2. }} In certain instances, conducting continued pre-training (\tapt or \pcp) on LMs with augmented data before preceding the FT can lead to substantial improvements. 
Specifically, this approach enhances the performance of prompt-based \ft (hard) from 46.7\% to 49.1\% on the SST-5 dataset and from 80.8\% to 85.9\% on the MPQA dataset. Notably, it boosts the performance of \cls-based \ft from 65.1\% to 75.3\% on the MPQA dataset, resulting in an approximate 6\% absolute value increase. 
These findings challenge the conclusions of prior research \cite{zhou-etal-2022-flipda} suggesting that DA techniques yield only minor gains.

%% file: table/dataset_statistics.tex
\begin{table*}[t!]
\begin{center}
\centering
\resizebox{\columnwidth}{!}{%
\begin{tabular}{lrrrrll}
\toprule
\bf Dataset & $|\mathcal{Y}|$ & $L$ & \bf \#Train & \bf \#Test & \bf Type & \bf Labels (classification tasks) \\
\midrule \rowcolor{Gray}
SST-5 & 5 & 18 & 8,544 & 2,210 & Sentiment & v. pos., positive, neutral, negative, v. neg. \\ 
MR & 2 & 20 & 8,662& 2,000 & Sentiment & positive, negative \\ \rowcolor{Gray}
CR & 2 & 19 & 1,775 & 2,000 & Sentiment & positive, negative \\ 
MPQA & 2 & 3 & 8,606 & 2,000 & Opinion Polarity & positive, negative \\ \rowcolor{Gray}
Subj & 2 & 23 & 8,000 & 2,000 & Subjectivity & subjective, objective \\ 
TREC & 6 & 10 & 5,452 & 500 & Question cls. & abbr., entity, description, human, loc., num.\\ \rowcolor{Gray}
STS-B & $\mathcal{R}$ & 11/11  & 5,749 & 1,500  & Sent. Similarity & - \\ 
\bottomrule
\end{tabular}
}
\end{center}
\vspace{-1.2em}
\caption{The datasets evaluated in this work. $|\mathcal{Y}|$: \# of classes for classification tasks (with one exception: STS-B is a real-valued regression task over the interval $[0, 5]$). $L$: average \# of words in input sentence(s).}
\vspace{-1em}
\label{table:datasets}
\end{table*}

%% file: table/template.tex
\newcommand\ttt[1]{\texttt{#1}}
\newcommand{\sent}{\ttt{<}$S_1$\ttt{>}}
\newcommand{\firstsent}{\ttt{<}$S_1$\ttt{>}}
\newcommand{\secondsent}{\ttt{<}$S_2$\ttt{>}}
\newcommand{\mask}{\texttt{[MASK]}}

\begin{table*}[!t]
\begin{center}
\centering
\resizebox{\textwidth}{!}{%
\begin{tabular}{lll}
\toprule
\tf{Task} & \tf{Template} & \tf{Label words}\\
\midrule
SST-5   &  {\sent} It was {\mask} . & v.positive: great, positive: good, neutral: okay,\\ \rowcolor{Gray}
        &                           & negative: bad, v.negative: terrible \\ 
MR      & {\sent} It was {\mask} .  & positive: great, negative: terrible\\ \rowcolor{Gray}
CR      & {\sent} It was {\mask} .  & positive: great, negative: terrible\\
MPQA    & {\sent} is {\mask} .      & positive: positive, negative: negative\\ \rowcolor{Gray}
Subj    & {\sent} This is {\mask} . & subjective: subjective, objective: objective \\ 
TREC    & {\mask} : {\sent}         & abbreviation: Expression, entity: Entity, description: Description \\ 
        &                           & human: Human, location: Location, numeric: Number \\ \rowcolor{Gray} 
STS-B & {\firstsent} {\mask} , {\secondsent} & $y_u$: Yes, $y_l$: No \\
\bottomrule
\end{tabular}
}
\end{center}
\vspace{-1.2em}
\caption{Templates and label words used for prompt-based \ft.}
\vspace{-1em}
\label{table:template}
\end{table*}

%% file: table/results.tex
\begin{table*}[th!]
\begin{center}
\centering
\resizebox{1.0\textwidth}{!}{%
\begin{tabular}{lccccccc}
\toprule
\tf{Dataset}                   & \tf{SST-5}            & \tf{MR}              & \tf{CR}               & \tf{MPQA}             & \tf{Subj}             &  \tf{TREC}       & \tf{STS-B}      \\
\tf{Evaluation Metrics}        & (acc)                 & (acc)                & (acc)                 & (acc)                 & (acc)                 & (acc)            & (Pear.)    \\
\midrule
Majority (full)                & 23.1       & 50.0         & 50.0         & 50.0         & 50.0         & 18.8          & -  \\
\midrule
\cls-based \ft                 & $41.7_{1.3}$          & $76.3_{3.2}$    \hl  & $79.5_{3.8}$          & $65.1_{12.6}$        & $91.7_{0.4}$          &  $80.3_{5.8}$   & $46.0_{16.3}$ \\
\tableindent + BT              & $40.8_{2.0}$\da       & $71.1_{5.7}$ \da      & $78.9_{3.2}$ \da      & $69.2_{4.3}$ \ua    & $91.0_{1.9}$\da   &  $83.1_{9.1}$ \ua  & $51.5_{22.6}$ \ua  \hl \\ 
\tableindent + \tapt           & $41.9_{2.2}$\ua\hl    & $76.1_{7.1}$ \da      & $85.3_{3.6}$ \ua\hl   & $75.3_{5.0}$ \ua\hl & $91.8_{1.2}$ \ua\hl   &  $83.8_{6.4}$\ua\hl& $41.9_{19.0}$ \da\\ 
\midrule
Prompt-based \ft (hard)        & $46.7_{1.5}$          & $86.2_{1.2}$         & $90.7_{0.8}$          & $80.8_{6.9}$          & $91.0_{1.1}$          & $84.7_{4.4}$       & $67.7_{8.1}$      \\
\tableindent + BT              & $45.4_{2.2}$ \da      & $85.5_{1.3}$ \da     & $91.1_{0.4}$  \ua     & $82.8_{5.1}$  \ua     & $91.3_{1.0}$  \ua     & $86.1_{4.3}$ \ua   & $66.3_{7.1}$ \da \\ 
\tableindent + \pcp            & $49.1_{1.5}$ \ua\hl   & $87.0_{1.4}$ \ua\hl  & $91.3_{0.9}$  \ua\hl  & $85.9_{1.9}$  \ua\hl  & $91.5_{1.3}$  \ua\hl  & $86.8_{3.9}$ \ua\hl& $70.1_{8.1}$ \ua \hl    \\
\midrule
Prompt-based \ft (soft)        & $48.0_{0.7}$          & $86.8_{1.4}$    \hl  & $90.8_{1.3}$          & $81.2_{6.8}$          & $90.3_{2.1}$          & $83.0_{3.0}$     & $63.7_{6.8}$      \\
\tableindent + BT              & $46.7_{0.9}$ \da      & $86.1_{1.4}$ \da     & $91.0_{0.9}$  \ua     & $82.9_{1.5}$   \ua\hl & $90.8_{1.0}$  \ua      & $85.8_{2.6}$ \ua  & $69.1_{8.4}$ \ua \\ 
\tableindent + \pcp            & $49.9_{1.2}$ \ua\hl   & $85.9_{1.4}$ \da     & $91.7_{1.2}$   \ua\hl & $84.6_{2.0}$  \ua     & $91.4_{1.5}$  \ua\hl  & $86.3_{2.3}$ \ua\hl & $69.6_{7.9}$ \ua\hl     \\
\bottomrule
\end{tabular}
}
\end{center}
\vspace{-1.2em}
\caption{
Test results using RoBERTa-large, where mean and standard deviation are reported over 5 seeds.
Green and red arrows indicate the positive/negative changes with respect to the \ft baselines that do not involve the back-translation.
The best performance on each dataset is highlighted in blue.
}
\vspace{-1em}
\label{table:few_shot}
\end{table*}

%% file: paper/conclusion.tex
In conclusion, our study challenges the notion of data augmentation's limited impact on FT LMs in \nlp tasks. We show that continued pre-training on augmented data can effectively improve model performance.